\documentclass[a4paper,twoside]{article}

\usepackage{epsfig}
\usepackage{subcaption}
\usepackage{calc}
\usepackage{amssymb}
\usepackage{amstext}
\usepackage{amsmath}
\usepackage{amsthm}
\usepackage{multicol}
\usepackage{pslatex}
\usepackage{apalike}
\usepackage{balance}\usepackage[bottom]{footmisc}
\usepackage{pifont}
\usepackage{comment}
\usepackage{multirow}

\usepackage[svgnames]{xcolor}

\usepackage{SCITEPRESS}    

\begin{document}

\title{Curriculum Learning for Compositional Visual Reasoning}

\author{\authorname{Wafa Aissa\sup{1,2}, Marin Ferecatu\sup{1} and Michel Crucianu\sup{1}}
\affiliation{\sup{1}Cedric laboratory, Conservatoire National des Arts et Métiers, Paris, France}
\affiliation{\sup{2}XXII Group, Paris, France}
\email{\{wafa.aissa, marin.ferecatu, michel.crucianu\}@lecnam.net}}


\abstract{Visual Question Answering (VQA) is a complex task requiring large datasets and expensive training. 
Neural Module Networks (NMN) first translate the question to a reasoning path, then follow that path to analyze the image and provide an answer. We propose an NMN method that relies on predefined cross-modal embeddings to ``warm start'' learning on the GQA dataset, then focus on Curriculum Learning (CL) as a way to improve training and make a better use of the data. 
Several difficulty criteria are employed for defining CL methods. We show that by an appropriate selection of the CL method the cost of training and the amount of training data can be greatly reduced, with a limited impact on the final VQA accuracy. Furthermore, we introduce intermediate losses during training and find that this allows to simplify the CL strategy. }

\onecolumn \maketitle \normalsize \setcounter{footnote}{0} \vfill

\section{\uppercase{Introduction}}

\label{sec:intro}

Visual Question Answering (VQA) consists in answering potentially complex questions regarding the content of images. Datasets like VQA2.0~\cite{balanced_vqa_v2} and GQA~\cite{Hudson2019} were put forward in support of this task. These datasets are very large, leading to expensive training. However, since they are built from collected real images with (possibly assisted) human labeling, they inevitably contain many biases. Integrated approaches like~\cite{mga,MIRTT} have the highest overall accuracies on these databases but are prone to taking bias-promoted ``shortcuts'', as shown \emph{e.g.}\ by their lower performance on out-of-distribution data~\cite{transfer}. Furthermore, integrated approaches lack transparency in the reasoning process, even though some limited explanations can be obtained by following the flow of attention.

Alternatively, Neural Module Networks (NMN) were introduced~\cite{LearnReason} with the aim to make the reasoning explicit. They were quite successful for visual reasoning on synthetic image datasets like CLEVR, where word grounding is comparatively easy and there is significant control over scene composition. NMNs are nevertheless hard to train on real images where grounding is difficult, attributes are more diverse and data biases are hard to control. To make learning more effective and less expensive, we rely on NMN but employ predefined cross-modal embeddings to ``warm start'' the training process on GQA, then explore Curriculum Learning to improve learning of such a complex task and reduce both the cost and the amount of data required.

Curriculum Learning (CL)~\cite{ELMAN199371,SovianyIRS22,TPAMI22CL} consists in learning the easier parts of the task first, rather than the entire task at once. However, adequate difficulty criteria are not easy to define. We show that by an appropriate selection of these criteria for VQA, the cost of training can be significantly reduced and less training data is required to reach a comparable level of accuracy. 

To summarize, the contributions of our work are three fold:
\begin{itemize}
\item First, we employ text and image object embeddings produced by a cross-modal transformer, with the goal of aligning multi-modal features to reinforce joint data patterns and thus help the learning process to achieve results faster.

\item Second, we propose several Curriculum Learning strategies to reduce both the training cost and the amount of data required for learning complex reasoning tasks.

\item Third, we define and employ intermediate module
losses (one per module) during training, using ground-truth labels generated from image graphs. The aim is to stabilize the learning and help the modules converge to the expected behavior defined by the modules' ground-truth and controlled by the local loss. 
\end{itemize}

The paper is organized as follows: the next section situates our proposals in the context of existing work on VQA and Curriculum learning, Sec.~\ref{sec:MCReas} describes the modular framework we employ and our use of cross-modal features. Then in Sec.~\ref{sec:cl} we define and motivate the CL strategies we propose. Evaluation results are presented and discussed in Sec.~\ref{sec:experiments}.

\section{\uppercase{Related Work}}
\label{sec:related-work}

We first review some recent work on visual reasoning for VQA. We then turn to the use of Curriculum learning for complex tasks, more specifically for VQA.

\subsection{Visual question answering}
VQA is usually addressed with either integrated cross-modal frameworks or compositional neural module networks.
 
\textbf{Cross-modality Transformers.} Transformer networks~\cite{transformer} have been widely applied to multiple language and vision tasks, and they have been recently adapted for reasoning problems such as VQA. Models like ViLBERT~\cite{vilbert}, VisualBERT~\cite{Visualbert} and LXMERT~\cite{lxmert} showcased good performances on the VQA datasets VQA2.0~\cite{balanced_vqa_v2} and GQA~\cite{Hudson2019}. 
These frameworks start by extracting the text and image features: word embeddings are obtained via a pretrained BERT~\cite{bert} model, while Faster R-CNN~\cite{FasterR-CNN} produces image region bounding boxes and corresponding visual features. Then, a cross-attention mechanism allows to align word embeddings and image features after training on a wide range of multi-modal tasks. 
One downside of integrated visual reasoning models is their lack of interpretability. Another drawback is their tendency to make ``shortcuts'' in reasoning, by learning the bias in the data as evidenced by their limited performance on the out-of-distribution data in GQA-OOD~\cite{transfer}. 
However, an effective cross-modal feature encoder can be obtained by discarding the final classification component from an integrated model. We employ here input features generated by an off-the-shelf large-scale cross-modal transformer encoder.

\textbf{Neural Module Networks (NMN).} To make the reasoning process more transparent and human-like, compositional NMNs~ \cite{LearnReason,pvr} perform multi-hop reasoning by decomposing a complex reasoning task into several easier sub-tasks. An NMN consists of a generator and an executor. The generator maps a question to a sequence of reasoning instructions (called a \emph{program}). The executor assigns each sub-task from this program to a neural module and passes the results to the next modules.
In~\cite{MMN} a meta-learning approach is employed in the NMN framework to improve the scalability and generalization of the resulting model. The generator decodes the question into a program whose sub-tasks are used to instantiate a meta module. The image features are extracted by a visual encoder implemented as a transformer network and a cross-attention layer mixes word embeddings and image features. 
While the combination of a generator and an executor in NMNs appears more complex than an integrated model, the ``hardwired'' reasoning process of an NMN is inherently transparent and has the potential to avoid part of the reasoning ``shortcuts'' caused by data bias. Interestingly, it was shown in~\cite{transfer} that by using the programs resulting from questions as additional supervision for the LXMERT integrated model allows to reduce sample complexity and improve performance on GQA-OOD. In our work, we aim to take advantage of both the transparency of NMN architectures and the quality of transformer-encoded representations by implementing a composable NMN over multimodal transformer vision and language features.

\subsection{Curriculum Learning}
\label{CL-pres}
Curriculum learning was introduced in~\cite{ELMAN199371} where the author shows that successful learning may depend on ``starting small'' by first learning a simple grammar with a recurrent network and then gradually learning more complex tasks such as relative clauses, number agreement, etc. CL was later applied to various machine learning tasks and recently adapted to textual question answering (QA) in~\cite{CLgen}. The authors use a sampling function that gives higher selection weights to simple 
QA pairs and then, as the training advances, it selects more complex QA pairs. A \emph{term frequency} selector and a \emph{grammar} selector assess the difficulty of the training examples. 
In~\cite{easy} CL is reframed as a self-paced learning (SPL) algorithm and the question loss is taken as the measure of difficulty. The authors implement several heuristics reminding of 
active learning in order to improve SPL performance.

\textbf{Curriculum Learning for VQA.} The definition of relevant difficulty criteria for VQA is challenging and this may explain why there is little work on the use of CL for VQA. The recent work in~\cite{CL-VQA} applies CL in a modular VQA context to the synthetic CLEVR dataset~\cite{CLEVR}. The base model is from~\cite{InfExec}, with an LSTM generator and generic residual blocks for the executor modules. The experiments 
were conducted on the executor alone, using as input the ground-truth programs directly. 
Several difficulty criteria were evaluated, including program length, answer hierarchy, and question loss. The results demonstrated that CL with a question loss difficulty criterion has a positive impact in a low data setting. 
However, the study in~\cite{CL-VQA} was focused on the CLEVR dataset~\cite{CLEVR} consisting of synthetic images of simple 3D objects, with a limited number of classes or attributes and reliable object detection. In our work, we employ the GQA dataset that is based on real-world images with many classes and several names for some of them, as well as more complex relations and more challenging object detection. We thus have to completely redefine the candidate CL strategies.

\section{\uppercase{Modular VQA framework}}

\begin{figure*}
  \includegraphics[width=\textwidth]{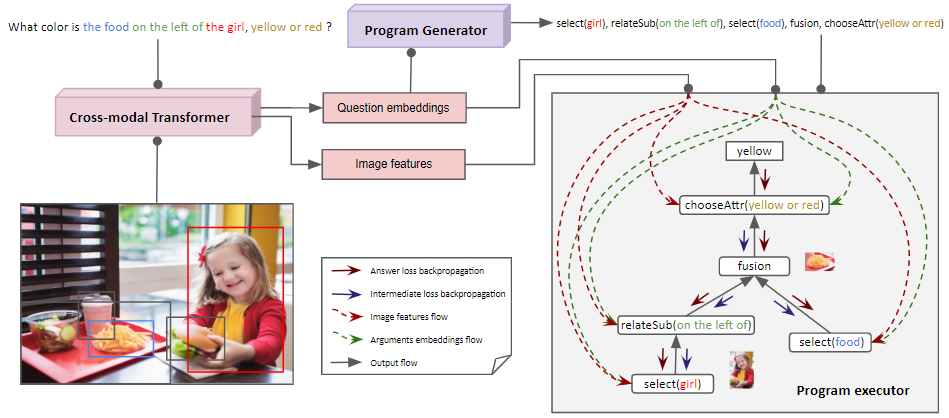}
  \caption{The proposed modular VQA framework: The (question, image) pair is used by a transformer model to generate aligned cross-modal embeddings for words and objects. These are used by a Program Generator module to produce a program (represented as a sequence of sub-task modules), which will be then applied by the Program Executor module to the image to answer the question. The proposed work focuses on improving the Program Executor by using several curriculum learning (CL) strategies.}
  \label{fig_vqa}
\end{figure*}
    
\label{sec:MCReas}

Our model takes as input a triplet composed of an image, a question and a program, and predicts an answer. 
We start by extracting aligned language and vision features for both the image and the question using a state-of-the-art cross-modal transformer. Then the program, which is a sequence of modules, is used to build the neural modules network that is executed on the image to answer the question (see Fig.~\ref{fig_vqa}). 
In the next subsections, we present the feature extraction process and describe the program executor.

\textbf{Cross-modal features.}
Compositional visual reasoning aims to perform logical and/or geometrical inferences involving several related objects in a complex scene. To achieve this, the reasoning modules are conditioned by text arguments. The visual and textual representations are vital to the reasoning process, therefore having good bounding box features and question text embeddings is crucial. 
To extract cross-modal language and vision representations we rely on LXMERT~\cite{lxmert}, a transformer model pretrained on multiple multi-modal tasks such as Masked Cross-Modality language models, masked object predictions, Cross-Modality Matching, and VQA. LXMERT showcases high accuracy on the training tasks, so we employ it as a feature extractor. It is worth mentioning that we only use the cross-modality encoder representations and discard the answer classification component.
More precisely, we freeze LXMERT weights and we pass the image~$I$ through the object-relationship encoder and the question~$Q$ through the language encoder. Then, the Cross-Modality Encoder aligns the representations to finally output the object bounding box features $v_j$ of each object $o_j$ in the image~$I$ and the embedding $h_i$ of each word $q_i$ in the question~$Q$.

\textbf{Neural Modules.} Our compositional reasoning model allows to perform complex reasoning tasks by decomposing them into easier sub-tasks. These sub-tasks are inspired by the human generic reasoning skills such as object detection, attribute identification, object relation recognition, object comparison, etc. 
We designed a library of modules where each module is responsible for performing a reasoning sub-task. The modules were designed to be intuitive and easily interpretable, each of them being implemented by a series of basic algorithmic operations such as dot products and MLPs. Modules can be categorized into three different groups based on their output type: attention, boolean and answer modules. For instance, an attention module such as \texttt{Select} is responsible of detecting an object bounding box by rendering an attention vector over the object bounding boxes contained in the image. Boolean modules such as \texttt{And} or \texttt{Or} make logical inferences and answer modules such as \texttt{QueryName} give a probability distribution over the answer vocabulary. Table~\ref{modules_small} shows an example from each module category. An exhaustive module list with definitions is provided in the appendix. 
\begin{table}[ht]
\caption{Sample module definitions. $S$:~softmax, $\sigma$:~sigmoid, $r$:~RELU, $\mathbf{W}_i$:~weight matrix, $\mathbf{a}$:~attention vector ($36\times1$), $\mathbf{V}$:~visual features ($768\times36$), $\mathbf{t}$:~text features ($768\times1$),
$\odot$: Hadamard product.} \vspace{1ex}
\label{modules_small}
{\small
\centering
\setlength\tabcolsep{1.5pt}
   \begin{tabular}{ | c | c | c | c | }
     \hline
      Name & Dependencies & Output & Definition  \\ \hline
      Select & $-$ & attention & $\begin{matrix}  
      \mathbf{x} = r(\mathbf{W}\,\mathbf{t}) \\ 
      \mathbf{Y} = r(\mathbf{W}\mathbf{V}) \\ 
      \mathbf{o} = S(\mathbf{W}(\mathbf{Y}^T \mathbf{x})) \\ \end{matrix}$ \\ \hline 
    RelateSub & [$\mathbf{a}$] & attention & 
     $\begin{matrix} 
    \mathbf{x} = r(\mathbf{W}\,\mathbf{t})\\
     \mathbf{Y} = r(\mathbf{W}\mathbf{V}) \\
     \mathbf{z} = S(\mathbf{W}(\mathbf{Y}^T \mathbf{x})) \\
     \mathbf{o} = S(\mathbf{W}(\mathbf{x} \odot \mathbf{y} \odot \mathbf{z}))
    \end{matrix}$ \\ \hline
      VerifyAttr & [$\mathbf{a}$] & boolean & 
     $\begin{matrix} 
     \mathbf{x} = r(\mathbf{W}\,\mathbf{t}) \\
     \mathbf{y} = r(\mathbf{W}(\mathbf{V}\,\mathbf{a}) \\
     \mathbf{o} = \sigma(\mathbf{W}(\mathbf{x} \odot \mathbf{y} ))
     \end{matrix}$\\
     \hline
      And & [$\mathbf{b}_1$,$\mathbf{b}_2$] & boolean & $\begin{matrix} \mathbf{o} = \mathbf{b}_1 \times \mathbf{b}_2 \\ \end{matrix}$ \\ \hline 
      ChooseAttr & [$\mathbf{a}$] & answer & $\begin{matrix} 
     \mathbf{x} = r(\mathbf{W}\,\mathbf{t}) \\
     \mathbf{y} = r(\mathbf{W}(\mathbf{V}\,\mathbf{a}) \\
     \mathbf{o} = S(\mathbf{W}(\mathbf{x} \odot \mathbf{y} ))
     \end{matrix}$ \\ \hline
      QueryName & [$\mathbf{a}$] & answer &  $\begin{matrix} 
      \mathbf{y} = r(\mathbf{W} (\mathbf{V}\,\mathbf{a})) \\ 
      \mathbf{o} = S(\mathbf{W}\,\mathbf{y})
      \end{matrix}$\\ \hline
   \end{tabular}
   }
\end{table}

\textbf{Modular network instantiation.} 
A program consists of a sequence of modules implemented as neural networks, as illustrated in Table~\ref{modules_small}. 
Each program is instantiated as a larger NMN following the sequence of program modules, where each module has dependencies $d_m$ to get information from the previous one, and arguments $a_m$ to condition its behavior. 
For example, the \texttt{FilterAttribute} module depends on the output of the \texttt{Select} module: it shifts the attention on the selected objects corresponding to the input text argument. 
The program executor is responsible of managing module dependencies by using a memory buffer to save the outputs that serve as inputs for the next modules.
The design insures that a module can have at most two dependencies. 

The generic modules require a textual argument $a_{m,i}$ to determine which facet of the module to use, for example, the \texttt{FilterAttribute} module can be called for multiple attribute categories such as color and size. For instance, to filter the red objects we use the argument word ``red'' and input its cross-modal embedding to the module.

\textbf{Module execution supervision.}
The executor network takes as input an image in the form of a list of object bounding boxes and their LXMERT embeddings. Every executor network ends with an answer module, the output of which is compared to the ground-truth to compute the output loss. To help modules converge faster to their expected behavior we also add the loss of each module to the output loss. 
The ground truth for each module is extracted from the image graphs given by the GQA dataset. Each image graph has $k$ assigned bounding boxes $bbox^*_{1...k}$ with their names, coordinates, attributes, and relations. Since we have two different intermediate module types, we define an intermediate loss for the attention modules and another for the boolean modules. 

\section{\uppercase{Curriculum learning for VQA}}
\label{sec:cl}

Our aim is to study CL for VQA and find a CL method that allows to significantly reduce training cost while making a better use of the data.

A CL method is usually defined by a difficulty criterion, a scheduler and a sampling function. 
The difficulty criterion allows to characterize the samples: training starts with the easiest samples, then progressively moves toward more difficult samples. Question loss was employed with some success as a difficulty criterion in~\cite{easy} for QA and in~\cite{CL-VQA} for VQA. 
However, computing question loss requires a first training iteration over all the training data. 
To make our own difficulty criteria, we assume that reasoning about a single object and its properties is simpler than examining the relations between several objects and comparing their attributes. The number of different objects in the question should then be a good indication of the complexity of reasoning and thus a relevant \emph{a priori} difficulty criterion for CL. 
Program length is another potentially relevant criterion that takes into account the flow of gradient in module networks corresponding to longer programs and is related to the previous criterion (more objects in the question require longer programs). 
Note that several criteria can be combined to define the increasing difficulty of the training samples in CL. When employing the number of objects as a primary criterion, we also evaluate its refinement based on program length: for each number of objects in the question, we start with the short programs, then continue with the medium length ones and end with the long programs.

The scheduler in CL decides when the curriculum should be updated. A simple solution is to employ a fixed sample size for each difficulty level. The evolution of the loss can be employed to adjust this size.  

The sampling function allows to modulate the selection of training samples within each difficulty level and works by assigning weights to all the examples. A relevant choice is to balance the occurrence probabilities of the different types of answer modules. Another criterion, used in boosting, is to privilege programs that lead to higher errors.

When the curriculum is updated, the new training sample has a higher level of difficulty than the previous ones. This does not mean that it subsumes the past samples and this is particularly true for the complex task of VQA. To avoid catastrophic forgetting, we add to the current sample (corresponding to the current difficulty level) a random selection from the past samples.

\section{\uppercase{Experiments}}
\label{sec:experiments}

\subsection{Experimental setup}

\textbf{GQA Dataset.} GQA~\cite{Hudson2019} features over 18M compositional questions and 113K real-world images. The reasoning steps of the questions are represented by functional programs. The questions and programs are generated by a question engine from the corresponding image graph. The image graphs are composed from ground-truth object bounding boxes together with their names, attributes, and relations. GQA is based on the Visual Genome dataset.

The GQA dataset has two versions: a balanced version (with a uniform distribution over the answers) and an unbalanced one. 
For the unbalanced version, the \texttt{train-all} split has over 14M examples and the \texttt{testdev-all} has 172,174 examples, while the balanced version has over 943,000 examples for \texttt{train}, 132,062 for \texttt{val} and a \texttt{testdev} of 12,578 examples.
To have a larger number of examples available for CL, we use the unbalanced GQA dataset. 
The experiments trained on the balanced dataset use the union 
of \texttt{train} and \texttt{val} as done in \cite{lxmert}, this combination also allows us to get over 1M examples when training on the balanced set.
We follow the recommendations of the dataset authors by evaluating the performance on the \texttt{test-dev} split instead of the \texttt{val} split when using the  object-based features because they were trained on some images from the \texttt{val} set \cite{butd}.

In the GQA dataset, each question/image pair in the \texttt{train}, \texttt{val}, and \texttt{testdev} sets is associated with a functional program. The programs use 124 distinct modules, some of which only correspond to very few questions. We consider that modules should only differ when they correspond to operations that are different. 
We thus group specific modules into more general ones. For example, modules like \texttt{ChooseHealthier} and \texttt{ChooseOlder} are grouped in a \texttt{ChooseAttribute} module. This results in only 32 modules, the list of which is presented in the appendix.

\textbf{Metrics.} Several metrics are employed to compare CL and standard learning. Since our focus is on reducing the cost of training, we measure the total number of example presentations during training (Comp.~cost). We also mention the maximal number of different examples seen during training (\#~examples), as different methods can make use of larger or smaller portions of the training set. While we focus on the cost of training, we nevertheless want to reach an accuracy that is close to the one obtained by standard learning, so we also report the accuracy of the predicted answers.

\subsection{Evaluated methods} 
\label{config}
When describing the different performed experiments, we use the following notations, which correspond to different algorithmic choices:\\
-- \textbf{Unbalanced}: We train 
on all the examples from the unbalanced GQA train split, we use the traditional random batch training strategy and the model sees all the data examples in every epoch. \\
-- \textbf{Balanced}: We train on the balanced version of the GQA dataset. At every epoch, the model is trained on all the balanced dataset examples. \\
-- \textbf{Random}: Instead of training on all the dataset examples, only 1M random examples from the unbalanced dataset are presented to the model at every iteration. \\ 
-- \textbf{CL}: The model is trained using Curriculum Learning and the filtering is driven by the number of objects in the programs. At every CL iteration the model sees 1M examples filtered from the unbalanced dataset by the curriculum sampler. The training needs 4 CL iterations to be complete, where each iteration has an increased difficulty given by the number of objects of its programs (ranging from 1 to 4).\\
-- \textbf{Length (L)}: The curriculum sampler filters the programs by their lengths for every number of objects, a CL-iteration is defined by a number of objects and a program length (short or medium or long). \\
-- \textbf{Weights (W)}: We use several sampling weights for the filtered programs: `uniform' indicates that the sampling is uniform, so the sampling with replacement results in a sample that is uniformly distributed over all the dataset. To make the answer modules distribution of the resulting sample more uniform we use the `answer module' weighting (denoted by W.a); this balances the answer modules appearances in the result sample so that the model equally sees all the defined answer modules.
The `modules loss' weighting (denoted by W.b) indicates that the example's weight is proportional to the sum of the average losses of the modules composing its program, to focus the model on hard examples. \\
-- \textbf{Pretrain (P)}: The model's parameters are initialized from a model trained using the \textbf{Random} variant described above.\\
-- \textbf{Repeat (R)}: We repeat the same CL-iteration twice. 

\subsection{Implementation details}
\label{section:implem}
We perform the experiments using the pre-processed GQA dataset programs and we focus on investigating the effect of several CL policies on the Program Executor module. Although our system also uses a transformer model as a generator to translate the question to its corresponding program, its task is relatively easy compared to the executor training and, similar to previous works~\cite{pvr,MMN}, we achieve near perfect translation results on the \texttt{testdev-all} set. 

We employ LXMERT as a feature extractor and freeze its weights. LXMERT image inputs are the object bounding boxes provided by a Faster R-CNN object detection model~\cite{butd}, where the number of bounding boxes per image is fixed to 36.
We feed the questions and their corresponding object bounding boxes to LXMERT; the encoding yields 36 object features and the question word embeddings for each question/image pair. The extracted features and embeddings have the size of the LXMERT hidden size, \emph{i.e.}\ 768.

During CL, we fix the sample size to 1M examples per CL iteration. 
Starting from the second CL iteration, 20\% of the training sample is sampled from the examples seen in the previous iteration.
Concerning the number of training examples, it is worth mentioning that we sample the training examples with replacement. Therefore, the number of \emph{distinct} examples seen by the executor is lower than the sample size.
To reduce the complexity of our proposed model, we allow weight sharing between compatible modules. For example, the \texttt{relateObj} and \texttt{relateSub} modules have similar structures and functionalities, both have visual and textual layers to project bounding box features and word embeddings in a multi-modal reasoning space (they also have an output layer to classify the answer). These two modules also have similar functionalities, \emph{i.e.}\ they both relate to an object given an anchor object and a relation but they have a different relation direction. Therefore, we share the weights between the visual layers and the textual layers of both modules respectively, but without sharing the output layers to guarantee transitional direction differences.
For training all the models we use SGD with a learning rate of 0.1 and a batch size of 1024.

\begin{table*}[t]
\caption{Results on \texttt{testdev-all} for several CL strategies. 
}
\label{results}
\centering
   \begin{tabular}{| l | c |  c | c | c | c | c |}
     \hline
     \multirow{2}{*}{Model} & \multicolumn{3}{c|}{CL configuration} & \multirow{2}{*}{Iterations} & Number of & \multirow{2}{*}{Accuracy} \\ \cline{2-4} 
     {}    & {weighting} & {pretraining} & iterations/level
     & {} & {examples ($\leq$)} & {} \\ \hline
     CL+W.a & answer & {$-$} & 1  & 4 & 4 M & 0.642 \\ \hline 
     CL+W.b &  losses & {$-$} & 1  & 4 & 4 M & 0.635 \\ \hline 
     CL+W.a+P &  answer & 2 iterations & 1  & $[2]+3$ & 5 M & 0.670 \\ \hline 
     CL+W.a+P+R &  answer & 2 iterations & 2 & $[2]+5$ & 7 M & \textbf{0.681} \\ \hline 
   \end{tabular}
\end{table*}

\begin{table}[b!]
\caption{Results on \texttt{testdev-all} with program length as a refinement for the CL difficulty measure. Computation cost is the number of seen examples per iteration times the number of iterations. 
}
\label{L}
\centering
\setlength\tabcolsep{3pt}
   \begin{tabular}{| l | c | c | c |}
     \hline
     Model & Comp.\ cost & \# examples & Accuracy \\ \hline
    CL+L &  11 & 11 M & 0.650 \\ \hline 
    CL+L+W.a & 12 & 12 M & 0.655 \\ \hline 
   \end{tabular}
\end{table}

\subsection{Evaluation results}

This section presents an analysis of the performance and the cost of our modular VQA framework with multiple CL training strategies, followed by a comparison with models not using CL to show the effectiveness of our proposed training approach. 

\textbf{Comparison of CL methods.} 
We start by a comparative analysis of the proposed CL strategies as described in Sec.~\ref{sec:cl}. Table~\ref{results} reports the performance of our model based on the different CL configurations detailed in Sec.~\ref{config}. The goal of CL is to make the training more effective and to achieve the highest accuracy while training for fewer iterations. Therefore, for each model are shown the number of iterations and training examples required to reach the highest accuracy.

From the results, it is clear that the `answer' weighting is the most effective weighting function. One can see this as a balancing of the answer modules presence over the training sample. 
The CL+W.a model (using the `answer' weighting) achieves higher accuracy results than the CL+W.b model (with the `losses' weighting), both reaching their top respective accuracies after 4 training iterations only. 
The `answer' weighting also yields better accuracy than the `uniform' weighting after the same number of training iterations. This is shown by comparing CL+L and CL+L+W.a in Table~\ref{L}. Moreover, the accuracy of CL+L+W.a continues to increase after the 11th iteration to achieve its top at iteration 12.
The superior performance of the `answer' weighting function in two different comparable settings makes us select this weighting for the rest of the experiments.

The refinement of the CL difficulty (or hardness) measure using the number of question objects (Length-CL difficulty measure) increases the CL+W.a top accuracy by 1\%, see the CL+L+W.a line in Table~\ref{L}. However, this improvement has a significant cost, as CL+L+W.a requires 12 training iterations (12M examples) unlike CL+W.a which only needs 4 iterations (4M examples). This reinforces the idea that with a more refined difficulty measure the model has more time to adjust to difficult examples, and its accuracy gradually increases to achieve a better top accuracy in a CL setting. But training on 12M examples is expensive since the overall dataset size is 14M examples. 
We thus decided to explore different options to obtain comparable results at a lower cost.

A promising finding was that pretraining the models for a few iterations with a randomly sampled 1M examples each leads to an accuracy increase of over $1.5\%$, as shown by the CL+W.a+P model which was pretrained for only 2 iterations.
This ``warms up'' the model to the modular aspect of our VQA framework, allowing it to be more general and effective before starting the CL. An interesting finding was that the model reached peak accuracy before iterating over the full CL configuration. 
The accuracy drop resulting after the 4th iteration may be explained by model overfitting on the questions with 4 objects. Indeed, in the GQA dataset these questions have a substantially unbalanced answer distribution.

A further finding is that repeating the same CL-iteration twice (as in CL+W.a+P+R) improves the top accuracy results by $1.1\%$, while only moderately increasing the number of iterations. This can be explained by the fact that doubling the number of training iterations helps the model better understand the structure of training data without augmenting the training data size.
As detailed in Sec.~\ref{section:implem}, when sampling with replacement we obtain a number of distinct examples that is slightly lower than the sample size, therefore the reported number of examples (\#~examples) is an upper bound of the \#~examples actually employed. 

As a general conclusion, we consider the CL+W.a+P+R model as the best modular VQA model that scores the best accuracy of 68.1\% after 7 training iterations using less than 7M distinct examples, \emph{i.e.}\ less than half of the training data. 

\textbf{Impact of CL}. 
We perform several experiments to assess the impact of the CL on our compositional visual reasoning framework. We do this by training our model without CL (Unbalanced, Balanced, and Random), then comparing the accuracy performance and the experiment cost in terms of computation cost and training data examples. In Table~\ref{ablation} we report the accuracy and cost results of the conducted experiments and compare them to the performance of our best CL model \textbf{CL+W.a+P+R}. 

The Unbalanced model (trained on the entire unbalanced training set of 14M) achieves the highest accuracy value of 70.2\%. 
This model also has the highest training cost among the evaluated models.

The Balanced model, trained on the balanced dataset for a large number of epochs, achieves lower results than the Unbalanced model. 
This is partly due to the fact that the balancing reduces not only the number of questions in the dataset, but also the diversity of the programs. Also, to the use of the unbalanced \texttt{testdev-all} for evaluation.

By comparing our best CL model (\textbf{CL+W.a+P+R}) to the models trained without CL (no-CL), we find very significant gains in terms of computational cost, \emph{e.g.}\ an 18-fold reduction compared to the top contender, the model trained on the Unbalanced dataset. 
The price to pay---a drop of only 2\% in accuracy---appears reasonable. 
The Random model, trained on randomly sampled 12M examples,  
performs almost as well as the Unbalanced model, an expected result since both models use a similar amount of distinct training examples (12M vs 14M). The Unbalanced model requires an almost 9 times more expensive training than Random, but the improvement in accuracy (70.2 \% vs.\ 69.4\%) hardly justifies it. However, the proposed CL model has an almost 2 times lower computational cost than Random, confirming the superiority of curriculum learning in this type of application.

\begin{table}[!h]
\caption{Comparaison of our CL model  (CL+W.a+P+R) with no-CL models (Unbalanced, Balanced, and Random) on the \texttt{testdev-all} set. }

\label{ablation}
\centering
\setlength\tabcolsep{3pt}
   \begin{tabular}{| l | c | c | c |}
     \hline
     Model & Comp.\ cost & \# examples & Accuracy \\ \hline
     Unbalanced  & $9 \times 14$ M & 14 M & \textbf{0.702} \\ \hline
     Balanced & $50 \times 1.4$ M & 1.4 M & 0.678 \\ \hline  
     Random & $12 \times 1$ M & $\leq 12$ M & 0.694 \\ \hline 
     CL+W.a+P+R & $\phantom{1}\mathbf{7} \times \mathbf{1}$ M & $< \mathbf{7}$ M & 0.681 \\ \hline 
   \end{tabular}
\end{table}

\section{\uppercase{Conclusion}}

In this work we present several Curriculum Learning (CL) strategies within a Neural Module Network (NMN) framework for Visual Question Answering (VQA). Our visual reasoning approach leverages a cross-modal Transformer encoder to extract aligned question/image features along with question programs to perform multi-step reasoning over the image and predict an answer. Our model employs an NMN architecture composed of multiple neural modules, each capable of performing a reasoning sub-task. 
We compare several CL strategies for VQA. Our model is evaluated on the GQA dataset and shows very interesting results in terms of computational cost reduction.
To drive the CL strategy, we introduce a difficulty measure based on the number of objects in the question and we achieve close accuracy results by training on a judiciously sampled 50\% of the training data, compared to an NMN model trained without CL on the entire training set.

\section*{ACKNOWLEDGEMENTS}
{We thank Souheil Hanoune for his insightful comments. This work was partly supported by the French Cifre fellowship  
2018/1601 granted by ANRT, and by XXII Group.}

\bibliographystyle{apalike}
{\small
\bibliography{example}}

\section*{\uppercase{Appendix}}

\balance

In this appendix, for reference purposes, we present the exhaustive list of modules together with their dependencies, types, and definitions (see Table~\ref{all}).
The column `output' represents the module type: Attention, Boolean, or Answer.

Attention modules produce an attention vector $\mathbf{a}$ where each element represents the relevance of the attended image object. Boolean modules make logical inferences and output a scalar representing the probability of the outcome. Answer modules give a probability distribution over the answer classes. 

As mentioned in Section~\ref{section:implem}, we used a weight sharing technique to reduce the number of the model parameters; this also allows the shared layers to have a better defined behavior and to be updated based on a larger number of training examples.
The overall principle is that a transfer is made only between some of the textual and visual layers, each module having a distinct output layer to guarantee its fine-tuning to the module's sub-task.

To decide what layers from which modules will share parameters, we assess the similarity between the modules by analyzing their functional and architectural properties. The former is derived from the module reasoning sub-task and the latter is derived from the module layer architectures.
We cannot exhaustively describe here all the performed analysis, but in the following, we exemplify our strategy by comparing some of the modules and explaining their inherent similarities and differences.   
The \texttt{Select} module detects a relevant bounding box (given the name of an object) and the \texttt{FilterAttr} detects a relevant bounding box given an attribute. Functionally, they both solve a detection problem but have different textual argument semantics. Architecturally, they both have the same layer structure: a textual layer, a visual layer, and an output layer. 
We decide to share the visual layer between these two modules but use different textual layers to respect the semantic differences between the textual arguments.

However, \texttt{FilterAttr} can share its textual layer with other modules having an attribute as a textual argument (\texttt{VerifyAttr}, \texttt{FilterNot}).

The \texttt{Same} and \texttt{Different} boolean modules assess whether or not two selected objects share the same characteristic (provided by the textual argument). The probability $p$ of two objects being similar is the opposite of them being different. Therefore, they share the same layers including the output layer and we use the relation $p(\mathrm{Different})=1-p(\mathrm{Same})$ to differentiate them.

The object relations modules such as \texttt{RelateSub} and \texttt{RelateObj} have similar functionalities and neural structures. They share their visual layers to get a common scene representation and they share the textual layer due to the semantic similarity of their arguments (a relation).

\begin{table*}[b]

\caption{Exhaustive module definitions. $S$:~softmax, $\sigma$:~sigmoid, $r$:~RELU, $\mathbf{W}$:~weight matrix, $\mathbf{a}$:~attention vector ($36\times1$), $\mathbf{b}$:~boolean scalar, $\mathbf{V}$:~visual features ($768\times36$), $\mathbf{t}$:~text features ($768\times1$),
$\odot$: Hadamard product, $[a\mathbin\Vert$b]: concatenation, $min$:~element-wise minimum.}
\centering
\label{all}
   \begin{tabular}{ | l | l | l | l | }
     \hline
     \textbf{Name} & \textbf{Dependencies} & \textbf{Output} & \textbf{Definition}  \\ \hline

     Select & $-$ & attention & 
    $\begin{matrix}  
      \mathbf{x} = r(\mathbf{W}\,\mathbf{t}), \mathbf{Y} = r(\mathbf{W}\mathbf{V}), \\ 
      \mathbf{o} = S(\mathbf{W}(\mathbf{Y}^T \mathbf{x}))  \end{matrix}$ \\
    \hline
    
     FilterAttr & [$\mathbf{a}$] & attention &  
    $\begin{matrix} 
    \mathbf{x} = r(\mathbf{W}\,\mathbf{t}), 
     \mathbf{Y} = r(\mathbf{W}\mathbf{V}) ,
    \mathbf{z} = S(\mathbf{W}(\mathbf{Y}^T \mathbf{x})), \\
    \mathbf{o} = \min(\mathbf{a},\mathbf{z})
    \end{matrix}$
      \\ \hline
      
       FilterNot & [$\mathbf{a}$] & attention &  $\begin{matrix} 
    \mathbf{x} = r(\mathbf{W}\,\mathbf{t}), 
     \mathbf{Y} = r(\mathbf{W}\mathbf{V}) ,
    \mathbf{z} = S(\mathbf{W}(\mathbf{Y}^T \mathbf{x})), \\
    \mathbf{o} = \min(\mathbf{a},1 - \mathbf{z})
    \end{matrix}$
    \\ \hline
     FilterPos & [$\mathbf{a}$] & attention & $\begin{matrix} 
    \mathbf{x} = r(\mathbf{W}\,\mathbf{t}), 
     \mathbf{Y} = r(\mathbf{W}\mathbf{V}) ,
    \mathbf{z} = S(\mathbf{W}(\mathbf{Y}^T \mathbf{x})), \\
    \mathbf{o} = \min(\mathbf{a},\mathbf{z})
    \end{matrix}$ \\ \hline
    
     RelateSub & [$\mathbf{a}$] & attention & 
     $\begin{matrix} 
    \mathbf{x} = r(\mathbf{W}\,\mathbf{t}),
     \mathbf{Y} = r(\mathbf{W}\mathbf{V}) ,
     \mathbf{z} = S(\mathbf{W}(\mathbf{Y}^T \mathbf{x})), \\
     \mathbf{o} = S(\mathbf{W}(\mathbf{x} \odot \mathbf{y} \odot \mathbf{z}))
    \end{matrix}$ \\ \hline
     
     RelateObj & [$\mathbf{a}$] & attention & $\begin{matrix} 
    \mathbf{x} = r(\mathbf{W}\,\mathbf{t}),
     \mathbf{Y} = r(\mathbf{W}\mathbf{V}) ,
     \mathbf{z} = S(\mathbf{W}(\mathbf{Y}^T \mathbf{x})), \\
     \mathbf{o} = S(\mathbf{W}(\mathbf{x} \odot \mathbf{y} \odot \mathbf{z}))
    \end{matrix}$
     \\ 
     \hline
     RelateAttr & [$\mathbf{a}$] & attention & $\begin{matrix} 
    \mathbf{x} = r(\mathbf{W}\,\mathbf{t}),
     \mathbf{Y} = r(\mathbf{W}\mathbf{V}) ,
     \mathbf{z} = S(\mathbf{W}(\mathbf{Y}^T \mathbf{x})), \\
     \mathbf{o} = S(\mathbf{W}(\mathbf{x} \odot \mathbf{y} \odot \mathbf{z}))
    \end{matrix}$ \\ \hline
     
     Fusion & [$\mathbf{a}_1$,$\mathbf{a}_2$] & attention & $ \mathbf{o} = \min(\mathbf{a}_1,\mathbf{a}_2)$ \\ 
     \hline
     \hline

     And & [$\mathbf{b}_1$,$\mathbf{b}_2$] & boolean & $\begin{matrix} \mathbf{o} = \mathbf{b}_1 \times \mathbf{b}_2 \\ \end{matrix}$ \\ \hline 
     Or & [$\mathbf{b}_1$,$\mathbf{b}_2$] & boolean & $\begin{matrix} \mathbf{o} = \mathbf{b}_1 + \mathbf{b}_2 - \mathbf{b}_1 \times \mathbf{b}_2 \end{matrix}$ \\
     \hline
     Same & [$\mathbf{a}_1$,$\mathbf{a}_2$] & boolean & 
     $\begin{matrix} 
     \mathbf{x} = r(\mathbf{W}\,\mathbf{t}),  \mathbf{y} = r(\mathbf{W}(\mathbf{V}\,\mathbf{a}_1)), \mathbf{z} = r(\mathbf{W}(\mathbf{V}\,\mathbf{a}_2)), \\ 
      \mathbf{o} = \sigma(\mathbf{W}(\mathbf{x} \odot \mathbf{y} \odot \mathbf{z})) 
    \\
     \end{matrix}$\\
     \hline
     SameAll & [$\mathbf{a}$] & boolean &
     $\begin{matrix} 
     \mathbf{x} = r(\mathbf{W}\,\mathbf{t}), 
     \mathbf{y} = r(\mathbf{W}(\mathbf{V}\,\mathbf{a})), \\
     \mathbf{o} = \sigma(\mathbf{W}(\mathbf{x} \odot \mathbf{y})) 
     \end{matrix}$ \\
     \hline
     Different & [$\mathbf{a}_1$,$\mathbf{a}_2$] & boolean & 
     $\begin{matrix} \mathbf{o} = 1 - same(\mathbf{a}_1,\mathbf{a}_2) \end{matrix}$
     \\
     \hline 
     
     DifferentAll & [$\mathbf{a}$] & boolean & $\begin{matrix} \mathbf{o} = 1 - same(\mathbf{a}) \end{matrix}$ \\
     \hline 
     
     Exist & [$\mathbf{a}$] & boolean & $ \begin{matrix}  \mathbf{o} = \sigma(\mathbf{W}([\mathbf{a} \mathbin\Vert \max(\mathbf{a}) \mathbin\Vert \min(\mathbf{a}) \mathbin\Vert mean(\mathbf{a})])) \\ 
     \end{matrix}$ \\
     \hline 
     
     VerifyRelSub & [$\mathbf{a}_1$,$\mathbf{a}_2$] & boolean & $\begin{matrix} 
      \mathbf{x} = r(\mathbf{W}\,\mathbf{t}), \mathbf{y} = r(\mathbf{W}(\mathbf{V}\,\mathbf{a}_1)), \mathbf{z} = r(\mathbf{W}(\mathbf{V}\,\mathbf{a}_2)), \\ 
      \mathbf{o} = \sigma(\mathbf{W}(\mathbf{x} \odot \mathbf{y} \odot \mathbf{z})) \end{matrix}$ \\
     \hline 
     VerifyRelObj & [$\mathbf{a}_1$,$\mathbf{a}_2$] & boolean & $\begin{matrix} 
      \mathbf{x} = r(\mathbf{W}\,\mathbf{t}), \mathbf{y} = r(\mathbf{W}(\mathbf{V}\,\mathbf{a}_1)), \mathbf{z} = r(\mathbf{W}(\mathbf{V}\,\mathbf{a}_2)), \\ 
      \mathbf{o} = \sigma(\mathbf{W}(\mathbf{x} \odot \mathbf{y} \odot \mathbf{z})) \end{matrix}$
     \\
     \hline
     VerifyAttr & [$\mathbf{a}$] & boolean & 
     $\begin{matrix} 
     \mathbf{x} = r(\mathbf{W}\,\mathbf{t}), 
     \mathbf{y} = r(\mathbf{W}(\mathbf{V}\,\mathbf{a}), \\
     \mathbf{o} = \sigma(\mathbf{W}(\mathbf{x} \odot \mathbf{y} ))
     \end{matrix}$\\
     \hline
     VerifyPos & [$\mathbf{a}$] & boolean & 
     $\begin{matrix} 
     \mathbf{x} = r(\mathbf{W}\,\mathbf{t}), 
     \mathbf{y} = r(\mathbf{W}(\mathbf{V}\,\mathbf{a}), \\
     \mathbf{o} = \sigma(\mathbf{W}(\mathbf{x} \odot \mathbf{y} ))
     \end{matrix}$ \\
     \hline
     \hline
     
     ChooseName & [$\mathbf{a}$] & answer & 
     $\begin{matrix} 
     \mathbf{x} = r(\mathbf{W}\,\mathbf{t}),
     \mathbf{y} = r(\mathbf{W}(\mathbf{V}\,\mathbf{a}), \\
     \mathbf{o} = S(\mathbf{W}(\mathbf{x} \odot \mathbf{y} ))
     \end{matrix}$
        \\ \hline
        
     ChooseAttr & [$\mathbf{a}$] & answer & $\begin{matrix} 
     \mathbf{x} = r(\mathbf{W}\,\mathbf{t}),
     \mathbf{y} = r(\mathbf{W}(\mathbf{V}\,\mathbf{a}), \\
     \mathbf{o} = S(\mathbf{W}(\mathbf{x} \odot \mathbf{y} ))
     \end{matrix}$ \\ \hline
     
     Compare & [$\mathbf{a}_1$,$\mathbf{a}_2$] & answer & 
     $\begin{matrix} 
     \mathbf{x} = r(\mathbf{W}\,\mathbf{t}),
     \mathbf{y} = r(\mathbf{W}(\mathbf{V}\,\mathbf{a}_1), 
     \mathbf{z} = r(\mathbf{W}(\mathbf{V}\,\mathbf{a}_2), \\
     \mathbf{o} = S(\mathbf{W}(\mathbf{x} \odot \mathbf{y} \odot \mathbf{z}))\\
     \end{matrix}$ \\ \hline
     
     ChoosePos & [$\mathbf{a}$] & answer & $\begin{matrix} 
     \mathbf{x} = r(\mathbf{W}\,\mathbf{t}),
     \mathbf{y} = r(\mathbf{W}(\mathbf{V}\,\mathbf{a}), \\
     \mathbf{o} = S(\mathbf{W}(\mathbf{x} \odot \mathbf{y} ))
     \end{matrix}$ \\ \hline
     
     ChooseRel & [$\mathbf{a}_1$,$\mathbf{a}_2$] & answer & $\begin{matrix} 
     \mathbf{x} = r(\mathbf{W}\,\mathbf{t}),
     \mathbf{y} = r(\mathbf{W}(\mathbf{V}\,\mathbf{a}_1), 
     \mathbf{z} = r(\mathbf{W}(\mathbf{V}\,\mathbf{a}_2), \\
     \mathbf{o} = S(\mathbf{W}(\mathbf{x} \odot \mathbf{y} \odot \mathbf{z} ))
     \end{matrix}$ \\ \hline
     
     Common & [$\mathbf{a}_1$,$\mathbf{a}_2$] & answer & 
     $\begin{matrix} 
     \mathbf{x} = r(\mathbf{W}(\mathbf{V}\,\mathbf{a}_1), 
     \mathbf{y} = r(\mathbf{W}(\mathbf{V}\,\mathbf{a}_2), \\
     \mathbf{o} = S(\mathbf{W}(\mathbf{x} \odot \mathbf{y} )) \\
     \end{matrix}$
     \\ \hline
     QueryName & [$\mathbf{a}$] & answer & 
      $\begin{matrix} 
      \mathbf{x} = r(\mathbf{W}(\mathbf{V}\,\mathbf{a}), 
      \mathbf{o} = S(\mathbf{W}(\mathbf{x}))
     \end{matrix}$ \\ \hline
     
     QueryAttr & [$\mathbf{a}$] & answer & $\begin{matrix} 
      \mathbf{x} = r(\mathbf{W}(\mathbf{V}\,\mathbf{a}), \\
      \mathbf{o} = S(\mathbf{W}(\mathbf{x})) \end{matrix}$ \\ \hline
      
     QueryPos & [$\mathbf{a}$] & answer & $\begin{matrix} 
      \mathbf{x} = r(\mathbf{W}(\mathbf{V}\,\mathbf{a}), \\
      \mathbf{o} = S(\mathbf{W}(\mathbf{x})) \end{matrix}$ \\ \hline
      
     AnswerLogic & [$\mathbf{b}$] & answer &  $\begin{matrix} 
     \mathbf{o_{yes} = \mathbf{b},  
     \mathbf{o}_{no} = 1-\mathbf{b}} 
     \end{matrix}$ \\ \hline 
     \end{tabular}
\end{table*}

\end{document}